\pdfoutput=1

\documentclass[11pt]{article}

\usepackage{ACL2023}

\usepackage{graphicx}

\usepackage{times}
\usepackage{latexsym}
\usepackage{adjustbox}
\usepackage[T1]{fontenc}

\usepackage[utf8]{inputenc}

\usepackage{microtype}

\usepackage{inconsolata}
\usepackage{amssymb}
\usepackage{booktabs}
\usepackage{amsmath}

\title{PCoQA: Persian Conversational Question Answering Dataset}

\author{Hamed Hematian Hemati$^\spadesuit$ \quad Atousa Toghyani$^\diamondsuit$ \quad Atena Souri$^\clubsuit$ \quad Sayed Hesam Alavian$^\spadesuit$ \\
 \textbf{Hossein Sameti$^\spadesuit$} \quad \textbf{Hamid Beigy$^\spadesuit$}\\ \\ 
    $^\spadesuit$AI Group, Computer Engineering Department, Sharif University of Technology \\
    $^\diamondsuit$Lorestan University \\
    $^\clubsuit$Tehran University
    }

\begin{document}
\maketitle
\begin{abstract}
Humans seek information regarding a specific topic through performing a conversation containing a series of questions and answers. In the pursuit of conversational question answering research, we introduce the PCoQA, the first \textbf{P}ersian \textbf{Co}nversational \textbf{Q}uestion \textbf{A}nswering dataset, a resource comprising  information-seeking dialogs encompassing a total of 9,026 contextually-driven questions. Each dialog involves a questioner, a responder, and a document from the Wikipedia; The questioner asks several inter-connected questions from the text and the responder provides a span of the document as the answer for each question. PCoQA is designed to present novel challenges compared to previous question answering datasets including having more open-ended non-factual answers, longer answers, and fewer lexical overlaps. This paper not only presents the comprehensive PCoQA dataset but also reports the performance of various benchmark models. Our models include baseline models and pre-trained models, which are leveraged to boost the performance of the model. The dataset and benchmarks are available at our Github page.\footnote{https://github.com/HamedHematian/PCoQA}
\end{abstract}

\section{Introduction}
In the realm of Question Answering (QA) systems, traditional approaches have largely focused on single-question scenarios, overlooking the dynamic nature of human information-seeking dialogs. However, to create more human-like and interactive QA systems, understanding context-dependent and evolving conversations is essential. To this end conversational question answering datasets have been introduced~\cite{DBLP:journals/tacl/ReddyCM19, DBLP:conf/emnlp/ChoiHIYYCLZ18, DBLP:conf/acl/CamposOSDCA20}. Here, we present PCoQA, an innovative and extensive \textbf{P}ersian \textbf{Co}nversational \textbf{Q}uestion \textbf{A}nswering dataset tailored explicitly for Question Answering in Context, drawing inspiration from two influential predecessors, CoQA~\cite{DBLP:journals/tacl/ReddyCM19} and QuAC~\cite{DBLP:conf/emnlp/ChoiHIYYCLZ18}. Our dataset contains 870 dialogs, 9,026 question-answer pairs, and corresponding documents, retrieved from the Wikipedia. We take initiatives from both prominent CoQA and QuAC datasets to build our dataset. To this end, like CoQA, both questioner and responder have access to the document in order to control the rate of unanswerable questions. Since questioner's accessibility to documents increases the odds of string matching and paraphrasing questions~\cite{DBLP:conf/emnlp/ChoiHIYYCLZ18}, two further measures are taken to diminish the phenomena. First, the questioner is informed to ask questions that do not contain lexical matching, and second, in the post-processing stage, questions that contain a high level of lexical overlap with the sentence containing the answer, are paraphrased to ensure the quality of the dataset.

Our dataset incorporates various linguistic phenomena related to conversations, including co-references to previous dialog turns, anaphora, and ellipsis. It introduces new challenges due to a higher presence of non-factual questions, resulting in longer answers. This characteristic is further compounded by the inclusion of abstract topics (since we don't value pages containing high number of entities over other pages) in our dataset, where documents often lack entities or noun phrases, and answers tend to be explanatory and lengthy.
Finally, we provide various benchmarks for the dataset, including baseline methods. Given that our dataset is approximately $\times10$ smaller than larger datasets like CoQA and QuAC, and data scarcity poses a challenge, we also explore the potential of enhancing model performance by pre-training it on other question-answering datasets. Our experiments exhibit the effectiveness of pre-training on boosting the performances.

The rest of the paper is structured as follows. We first describe the previous datasets and methods in Section \ref{sec:related-works}. Subsequently, we provide the details of building the dataset in Section \ref{sec:dataset}, comprising document selection, data annotation, post-processing, dataset validation, dataset analysis, and splitting. Lastly, in Section \ref{sec:experiments}, our tested models, experiments, and results are reported.

\section{Related Works}
\label{sec:related-works}
\vspace{-\parskip}
Multiple datasets have been introduced for the task of question answering \cite{DBLP:conf/emnlp/RajpurkarZLL16, DBLP:conf/rep4nlp/TrischlerWYHSBS17, DBLP:journals/corr/DunnSHGCC17, DBLP:journals/tacl/KwiatkowskiPRCP19}. The field of conversational question answering aims to extend systems' capabilities in answering questions within the conversational domain. Multiple datasets for this task have been proposed in English \cite{DBLP:conf/emnlp/ChoiHIYYCLZ18, DBLP:journals/tacl/ReddyCM19, DBLP:conf/acl/CamposOSDCA20}. Unlike QA domain, where multiple datasets in various languages are available \cite{DBLP:journals/corr/abs-1912-05200, DBLP:journals/corr/abs-1806-00920, DBLP:journals/corr/abs-2112-13634}, attempts to build datasets for CQA in non-English languages have been limited \cite{DBLP:conf/lrec/OtegiGCSA20}. \citet{DBLP:conf/lrec/OtegiGCSA20} constructed a CQA dataset in the Basque language.
\newline
Multiple methods have been proposed to effectively model the history in CQA. \citet{DBLP:conf/sigir/Qu0QCZI19} proposes marking history answers in the embedding layer, and \citet{DBLP:conf/cikm/QuYQZCCI19} extends this work by considering the order of histories. Another line of research utilizes question rewriting \cite{DBLP:conf/wsdm/VakulenkoLTA21} to address the problem. \citet{DBLP:conf/acl/KimKPK20} employs consistency training to mitigate the error propagation problem in rewritten questions, while \citet{DBLP:conf/emnlp/ChenZFFRM22} uses reinforcement learning to rewrite questions based on feedback from a question-answering module. Despite these significant efforts, a notable issue in most of the mentioned research is the use of ground truth answers as part of the modeling process. \citet{robust} re-implements the works of \citet{DBLP:conf/sigir/Qu0QCZI19, DBLP:conf/cikm/QuYQZCCI19} without utilizing the gold answers of history and reports significantly lower performance. \citet{DBLP:conf/lrec/OtegiGCSA20} adopts pre-training on other resources to alleviate the impact of low-resource data.


\section{Dataset Construction} \label{sec:dataset}
This section describes the process of constructing PCoQA dataset. A dialog sample of the dataset is depicted in Figure~\ref{fig:pcoqa_sample} whose title is Pride \& Prejudice.
\subsection{Document Collection}
The documents in our dataset are based on Wikipedia articles. In line with previous question-answering datasets in Persian~\cite{DBLP:journals/csl/DarvishiSAM23, PersianQA}, we have chosen Wikipedia as the primary source for obtaining documents. To build our documents, we have taken a different approach compared to CoQA, which selects the initial portion of each article as the final document~\cite{DBLP:journals/tacl/ReddyCM19}. Wikipedia articles typically begin with an abstract that provides general information about the article's topic, and subsequent sections delve into specific details. While the abstract is essential for constructing final documents, the finer-grained information in the subsequent sections should not be overlooked.
To address this concern and ensure diversity among documents with consistent contexts, we have devised a unique process for building our final documents.
Initially, we select two bounds for the minimum and maximum document lengths, denoted as $D_m$ and $D_M$ respectively. $D_m$ is set to ensure that all documents contain a minimum context necessary for a meaningful dialog, while $D_M$ prevents excessively long documents that can challenge current network modeling capabilities, as transformers consist our main models and they receive a limited length as input. In practice we set $D_m=100$ and $D_M=1000$.
In our approach, we differentiate between the abstract and other sections, which we represent as $A$ and $S_i$, respectively, with $i$ being the section number. The lengths of $A$ and $S_i$ are denoted as $L_{A}$ and $L_{S_i}$, respectively. To ensure consistency in the context of the final documents, we appoint a human annotator as the document provider. The role of the document provider is crucial in curating documents that align with the dataset's requirements and maintain contextual coherence.
First, a Wikipedia page is chosen on random. Unlike \citet{DBLP:journals/tacl/ReddyCM19}, we don't consider any pre-condition, like a good number of entities, for the selected pages. This is because we want to maximize diversity. For instance, it's obvious that pages regarding abstract phenomena contains a few entities whereas pages regarding individuals and geographical locations contain significant amount of entities. Next, It is decided to whether select the document from the $A$ or a random $S_i$-s:

\begin{itemize}
    \item 
    If $A$ is selected as the beginning of the document: If $L_A + \sum_i L_{S_i} \leq D_m$, meaning that the length of page is below the minimum constraint, the document is discarded and if $A \geq D_M$, the $A$ is tailored to ensure the maximum length constraint. If $A$ is chosen and $D_m \leq L_A \leq D_M$, $A$ is selected as the document. To encourage diversity, the document provider is allowed to append $S_i$-s to the $A$ in order to elongate the document such that the maximum length constraint is preserved; these $S_i$-s are selected in a way so that they are semantically consistent with each other and $A$.
    \item \label{Z}
    If $S_j$ is selected as the beginning of the document: the process follows a similar pattern as previously described. However, in this case, the document begins with $S_j$ and is subsequently extended with potentially semantically consistent sections, as determined by the document provider.
\end{itemize}

To illustrate the process, an example involving the Wikipedia page for "Canada" is shown in Figure \ref{fig:canada}. We begin by selecting $S_{11}$, which corresponds to the "Education System" section of the page. Since the length of this section, $L_{S_{11}}$, is within the bounds defined by $D_m$, the document provider proceeds to choose the next two sections, namely "Economy" and "Culture" These sections are semantically consistent with the subject of "Education System". The final document is then composed by concatenating these three selected sections.

It is important to emphasize the pivotal role of the document provider in this process. The document provider must carefully oversee the content of each $S_i$ to ensure consistency. For instance, if $S_j$ is selected as the beginning of the document and it contains a co-reference to a previous section or some of its content is vague due to lack of previous context, $S_j$ should be omitted from the selection, and the process should proceed with the next suitable section.

\subsection{Dataset Annotation}
To establish dialogs, each document is assigned to to a questioner and a responder, both of whom having access to the title and text of the document. At the turn of $k$, questioner asks a question of $q_k$ and the responder returns $a_k$, a span of the document as the answer. The dialog is continued until the questioner stops the conversation. The questioners are informed that they should start the conversation with general information and continue it to specific subjects, to match the same process of human information seeking in real world. To be specific, they're strictly told that they should not ask about specific concepts regrading a topic unless they're informed about that concept in previous dialog turns. Additionally, questioners are informed to change their questions if their questions exhibit a substantial overlap with the potential answer.

\subsection{Post-Processing}
While our dataset is designed to provide questioners with access to documents, it is possible that string-matching questions may arise~\cite{DBLP:conf/emnlp/ChoiHIYYCLZ18}, despite our efforts to guide questioners to avoid such issues. Previous studies have indicated that questions exhibiting high similarity to the sentence containing the answer have a greater likelihood of being answered correctly~\cite{DBLP:conf/emnlp/SugawaraISA18}. To ensure the dataset's quality, we have identified these questions and had them rewritten to reduce lexical overlap between the rewritten question and the sentence that contains the corresponding answer. Each question that shares at least one similar word with the answer-containing sentence is subjected to this rewriting process. A question is rewritten in one of three ways:
\begin{itemize}
    \itemsep-.1em 
    \item 
    Words were removed due to ellipsis
    \item
    Words were replaced by their synonyms
    \item
    Words were replaced by their co-references
\end{itemize}

We quantified the similarity using the formula $\text{similarity}=\frac{|~overlap~|}{|~question\:words~|}$ where $overlap$ is the set of shared words between the question and the sentence containing the answer. Before the rewriting process, the similarity was measured at 14.2, and after rewriting, the similarity was reduced to 11.8.

\subsection{Dataset Validation} 
Following previous research~\citet{DBLP:conf/emnlp/ChoiHIYYCLZ18, DBLP:conf/emnlp/RajpurkarZLL16}, multiple annotations are provided for each question in Dev/Test set. This is due to the fact that each question can have multiple answers; Therefore, it is indispensable to obtain accurate and unbiased scores for evaluation. These annotations are tagged by annotators other than responders.
In line with previous research \cite{DBLP:conf/emnlp/ChoiHIYYCLZ18, DBLP:conf/emnlp/RajpurkarZLL16}, multiple annotations are assigned to each question in the Dev/Test set. This practice is essential because a single question may have multiple valid answers. It ensures the acquisition of accurate and unbiased scores for evaluation purposes. Notably, these annotations are provided by annotators who are distinct from the responders. We report the scores of the responders' answers in Table \ref{table:res}.

\subsection{Dataset Analysis}
Key statistics for the PCoQA dataset are presented in Table \ref{table:stat}, along with a comparison to similar datasets such as CoQA and QuAC. In the table, the expression $X / Y$ represents the average quantity of $X$ per unit of $Y$. Notably, PCoQA answers are longer, reflecting the prevalence of non-factual questions in our dataset. Additionally, our documents are longer than those in CoQA and QuAC, necessitating the use of transformers with larger input sizes, as standard transformers have limited input capacities. Furthermore, our dataset features a higher number of questions per dialog compared to QuAC, underscoring the importance of effective history representation.

\begin{table}[h!]
\centering
\begin{adjustbox}{width=.45\textwidth}
\begin{tabular}{cccc} \toprule
 & \multicolumn{1}{c}{PCoQA} & \multicolumn{1}{c}{CoQA} & \multicolumn{1}{c}{QuAC} \\ 
  \midrule
    documents & 870 & 8,399 & 11,568 \\
    questions & 9,026 & 127,000 & 86,568 \\
    tokens, words / document & 505.4 & 271.0 & 401.0 \\
    tokens, words / question & 7.0 & 5.5 & 6.5 \\
    tokens, words / answer & 18.6 & 2.7 & 14.6 \\
    questions / dialog & 10.4 & 15.2 & 7.2 \\
    unanswerable rate & 15.7 & 1.3 & 20.2 \\
     
  \bottomrule
\end{tabular}
\end{adjustbox}
\caption{Statistics of the PCoQA Dataset}
\label{table:stat}
\end{table}

\subsection{Splitting}
The dataset is randomly divided into Train, Dev, and Test sets with the ratio of 70/15/15.

\section{Experiments}\label{sec:experiments}
In this section, we describe the adopted evaluation metrics, methods, and the results of applying these methods to the PCoQA dataset.

\subsection{Evaluation Metrics}
Exact Matching (EM) is the ratio of questions for which the model has answered correctly. Following~\citet{DBLP:conf/emnlp/ChoiHIYYCLZ18}, three additional metrics of F1, HEQ-Q, and HEQ-D are considered in this paper. F1 indicates the degree of overlap between the predicted answer and the gold answer, and HEQ-Q and HEQ-D are the ratio of questions and dialogs for which the model outperforms the human respectively~\cite{DBLP:conf/emnlp/ChoiHIYYCLZ18}. While HEQ-D is a stringent metric that requires the model to outperform humans for every question within a dialog to earn a point, it may be overly strict in some cases. While HEQ-D is a stringent metric that requires the model to outperform humans for every question within a dialog to earn a point, it may be overly strict in some cases. To address this, we introduce another metric, called HEQ-M. HEQ-M quantifies the number of dialogs for which the model achieves a better overall performance compared to human performance on average. Additionally, we analyze the F1 score for each dialog turn to gain insights into the model's performance at different turns of the conversation.

\subsection{Importance of History}
\begin{figure}[htb]
    \centering
		\includegraphics[width=.9\columnwidth]{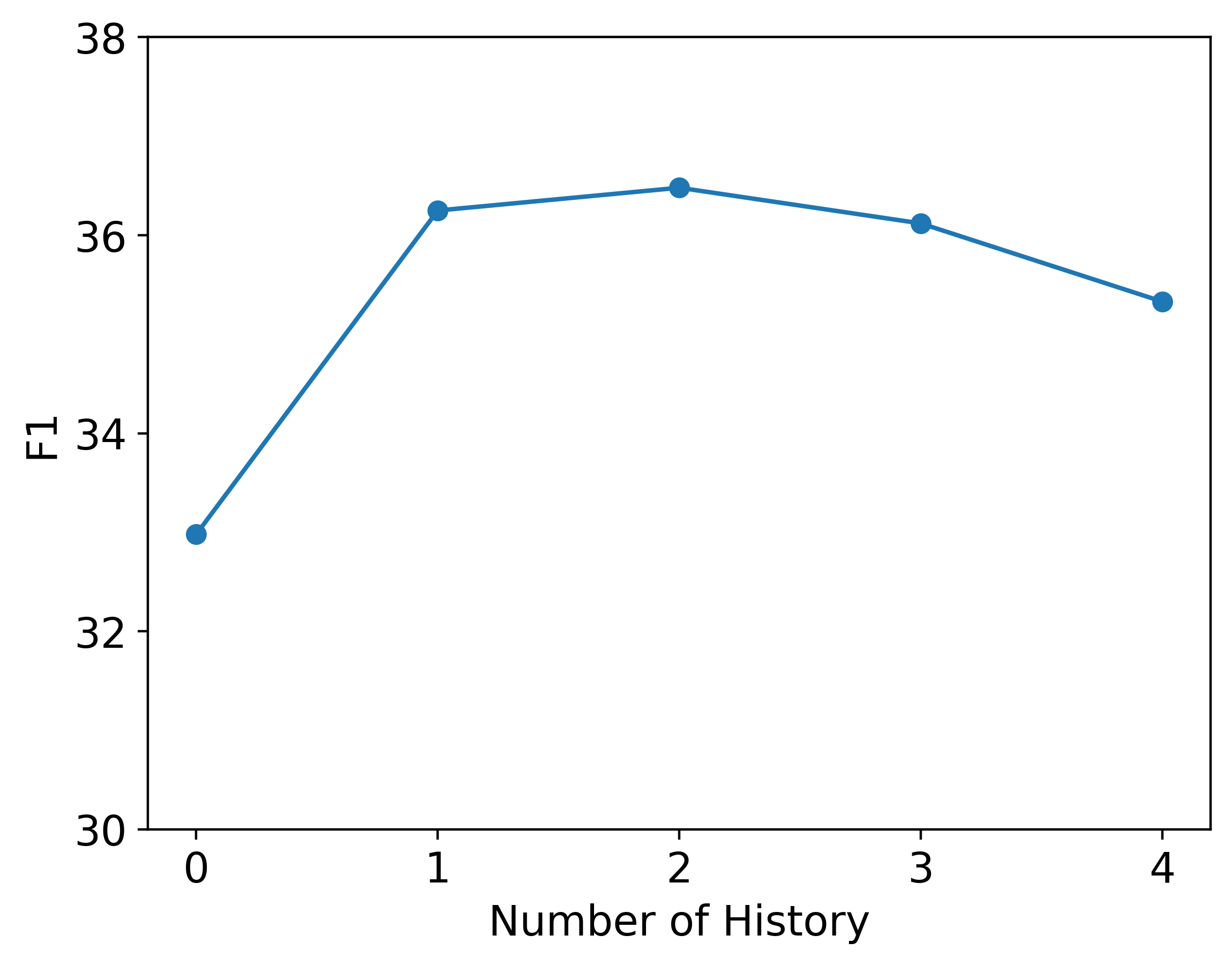}
		\caption
		{
            Effect of history number on performance
		}
		\label{fig:XX}
\end{figure}

In this section, we explore the impact of history on model performance. Figure \ref{fig:XX} illustrates the performance variation of the model concerning the inclusion of a different number of history questions. Notably, excluding the history questions results in a sharp drop in the model's performance. The best performance is achieved when using $2$ history questions. However, including more than $2$ history questions gradually leads to a decline in performance. This suggests that histories with distances over $2$ are irrelevant and don't introduce new information on average, and their inclusion induces some noise in the model. Thus, we perform the rest of our experiments with $2$ history turns.

\begin{table*}[!h]
\centering
\begin{tabular}{cccccc} \toprule
Model & \multicolumn{1}{c}{EM} & \multicolumn{1}{c}{F1} & \multicolumn{1}{c}{HEQ-Q} & \multicolumn{1}{c}{HEQ-M} &
\multicolumn{1}{c}{HEQ-D}\\ 
  \midrule
  ParsBERT & 21.82 & 37.06 & 30.70 & 0.0 & 0.0 \\
  XLM-Roberta & 30.47 & 47.78 & 39.51 & 2.45 & 1.63\\
  \midrule
  ParSQuAD + ParsBERT & 21.74 & 40.48 & 31.95 & 0.8 & 0.0  \\
  QuAC + XLM-Roberta & 32.81 & 51.66 & 43.10 & \textbf{3.27} & \textbf{1.63} \\
  ParSQuAD + XLM-Roberta & \textbf{35.93} & \textbf{53.75} & \textbf{46.21} & 1.63 & 0.8 \\
  \midrule
  Human & 85.50 & 86.97 & - & - & - \\
  \bottomrule
\end{tabular}
\caption{Results of different models across metrics}
\label{table:res}
\end{table*}

\subsection{Methods}
Our experimented methods can be categorized into two main groups: baseline methods and methods based on pre-training. Our experimental framework is built upon two base transformer model \cite{DBLP:conf/nips/VaswaniSPUJGKP17}: ParsBERT \cite{DBLP:journals/npl/FarahaniGFM21}, a Persian equivalent of BERT \cite{DBLP:conf/naacl/DevlinCLT19}, and XLM-Roberta \cite{DBLP:conf/acl/ConneauKGCWGGOZ20}. These base models serve as the foundation for our methodology. In our implementation, each model takes the concatenated question and previous history questions as the first input and the document as the second input, which is then fed into the transformer.
\paragraph{Baseline Methods}
ParsBERT and XLM-Roberta are fine-tuned on PCoQA, constituting our baseline methods.
\paragraph{Pre-Trained Methods}
ParSQuAD + ParsBERT denotes pre-training ParsBERT on ParSQuAD~\cite{abadani2021parsquad}, a translated dataset of SQUAD~\cite{DBLP:conf/acl/RajpurkarJL18} to Farsi, and then fine-tuning it on PCoQA using history concatenation. Similarly, ParSQuAD + XLM-Roberta denotes pre-training XLM-Roberta on ParSQuAD and then fine-tuning it on PCoQA using history concatenation. Lastly, QuAC + XLM-Roberta represents pre-training XLM-Roberta on QuAC and subsequently fine-tuning it on PCoQA using history concatenation.

\subsection{Results}
The results of our experiments are presented in Table \ref{table:res}, where we evaluate the performance across all metrics. It's evident that XLM-Roberta outperforms ParsBERT, highlighting the superior capabilities of XLM-Roberta. Moreover, our experiments demonstrate the effectiveness of pre-training techniques. The highest scores are achieved by XLM-Roberta when pre-trained on ParSQuAD. However, even with this strong performance, there remains a substantial gap between our models' scores and those of human responders. Notably, this gap is especially pronounced in the EM score. We observe that humans tend to provide complete answers when they know the answer, as evidenced by the nearly equal F1 and EM scores. In contrast, our models exhibit a significant disparity between F1 and EM scores, suggesting that they may struggle to provide complete answers, even when they partially address the questions.

\begin{figure*}[h!]
	\begin{center}
		\includegraphics[width=1.0\textwidth]{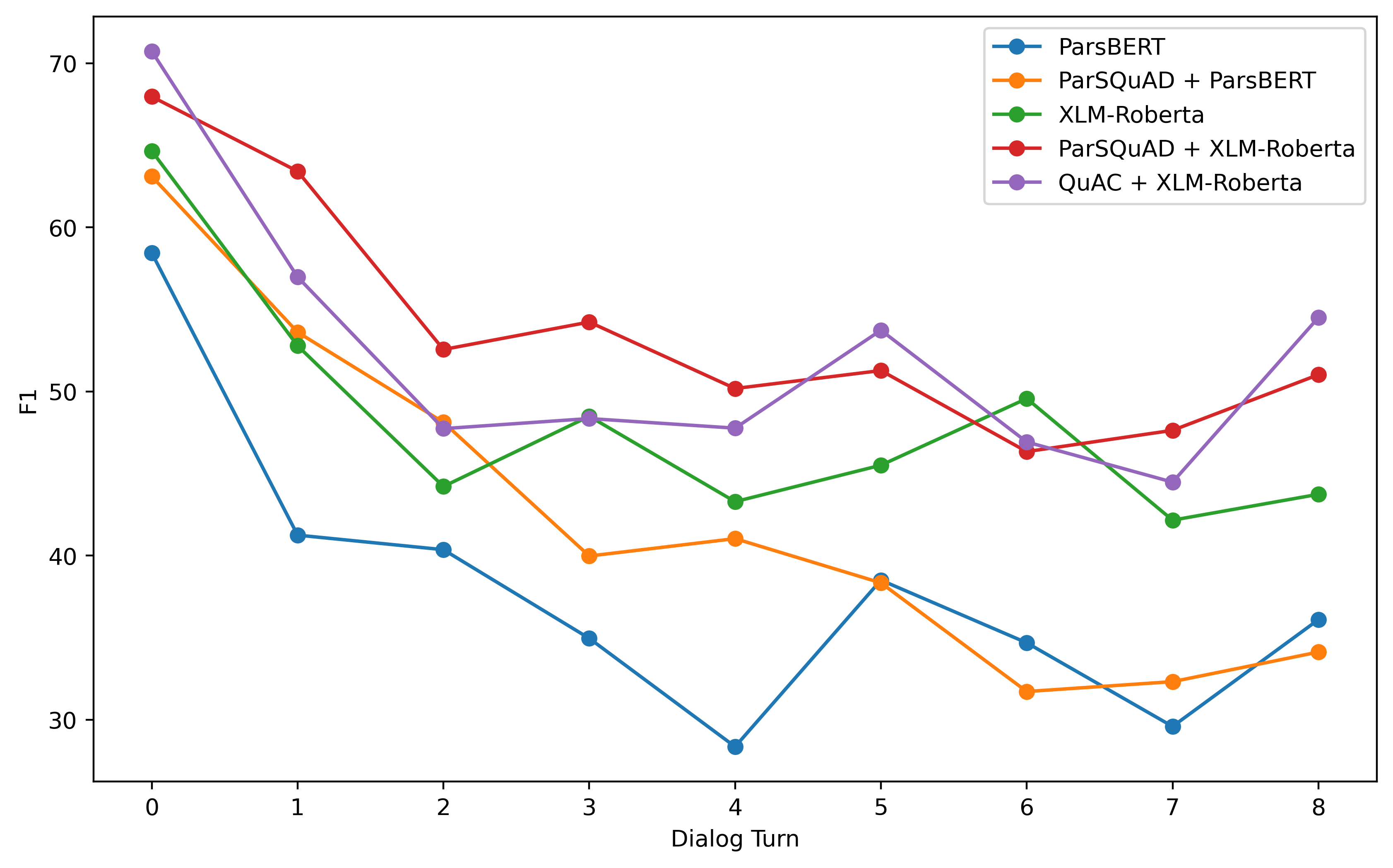}
		\caption
		{
            F1 scores of each dialog turn across different models
		}
		\label{fig:turn_f1}
	\end{center}
\end{figure*}

\subsection{Pre-Training Effect}
We observe that when using XLM-Roberta, ParSQuAD gives better results compared to QuAC. This observation is notable because QuAC, in contrast to ParSQuAD, is a conversational dataset. It suggests that XLM-Roberta may encounter challenges when jointly modeling English and Persian. Consequently, pre-training on ParSQuAD, which is in Persian like PCoQA, outperforms pre-training on QuAC.
Furthermore, we find that pre-training on QuAC improves performance on metrics like HEQ-M and HEQ-D, indicating that it imparts valuable conversational information, specifically the dependency among questions, to our model. This observation is reinforced by our findings in Figure \ref{fig:turn_f1}, where we observe that, ar initial turns, ParSQuAD+XLM-Roberta outperforms other QuAC+XLM-Roberta. However, as the conversation progresses, QuAC+XLM-Roberta achieves performance on par or better than ParSQuAD+XLM-Roberta, further underscoring the value of conversational pre-training.
A similar pattern can be observed when examining the performance of ParSQuAD+ParsBert compared to ParsBert, as depicted in Figure \ref{fig:turn_f1}. Initially, the performance of ParSQuAD+ParsBert is superior, but as the conversation evolves, the performances of ParSQuAD+ParsBert and ParsBert become comparable, suggesting that pre-training on ParSQuAD does not effectively capture conversational information.

\section{Conclusion}
In this paper, we introduce PCoQA, the first Persian conversational question-answering dataset, constructed using Wikipedia pages. Distinguishing itself from some previous works, our dataset emphasizes diversity. We establish ParsBERT and XLM-Roberta as our baseline models. Due to our dataset's size limitations compared to current English datasets, we explore pre-training on existing datasets, ParSQuAD and QuAC, and found pre-training effective. While ParSQuAD pre-training generally yields better results, it falls short in effectively transferring conversational information to the target task.
For future work, we suggest approaching conversational question-answering dataset construction through synthetic or semi-automatic methods to minimize artifacts. Additionally, it would be valuable to evaluate previous methods, excluding history answers, on the PCoQA dataset and compare the results with our findings.

\bibliography{custom}
\bibliographystyle{acl_natbib}

\clearpage
\onecolumn 
\section{Appendix}
\renewcommand{\thefigure}{A\arabic{figure}}
\setcounter{figure}{0}
\begin{figure*}[h!]
	\begin{center}
		\includegraphics[width=.6\textwidth]{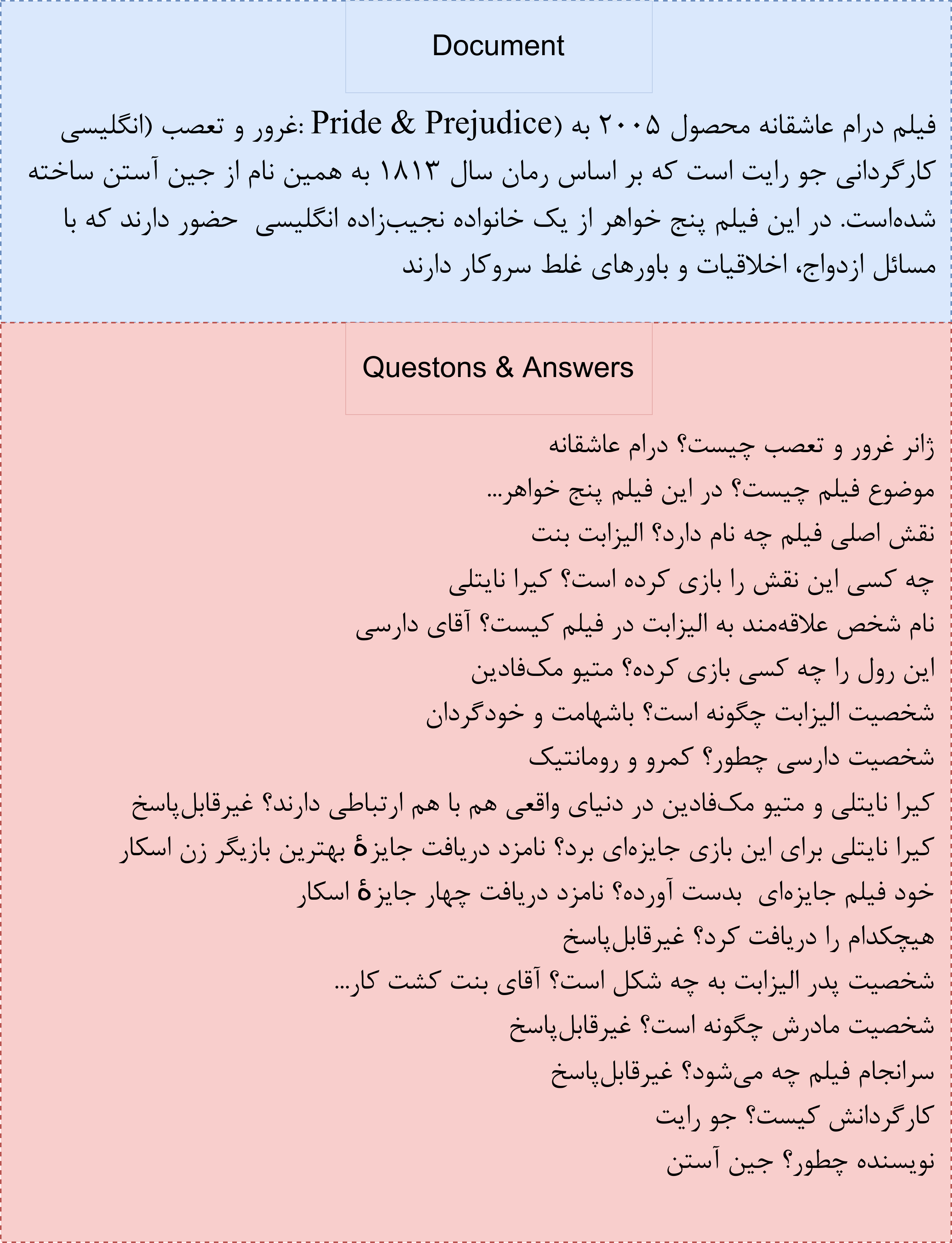}
		\caption
		{
            A document and its corresponding questions/answers dialog
		}
		\label{fig:pcoqa_sample}
	\end{center}
\end{figure*}

\begin{figure*}[h!]
	\begin{center}
		\includegraphics[width=.9\textwidth]{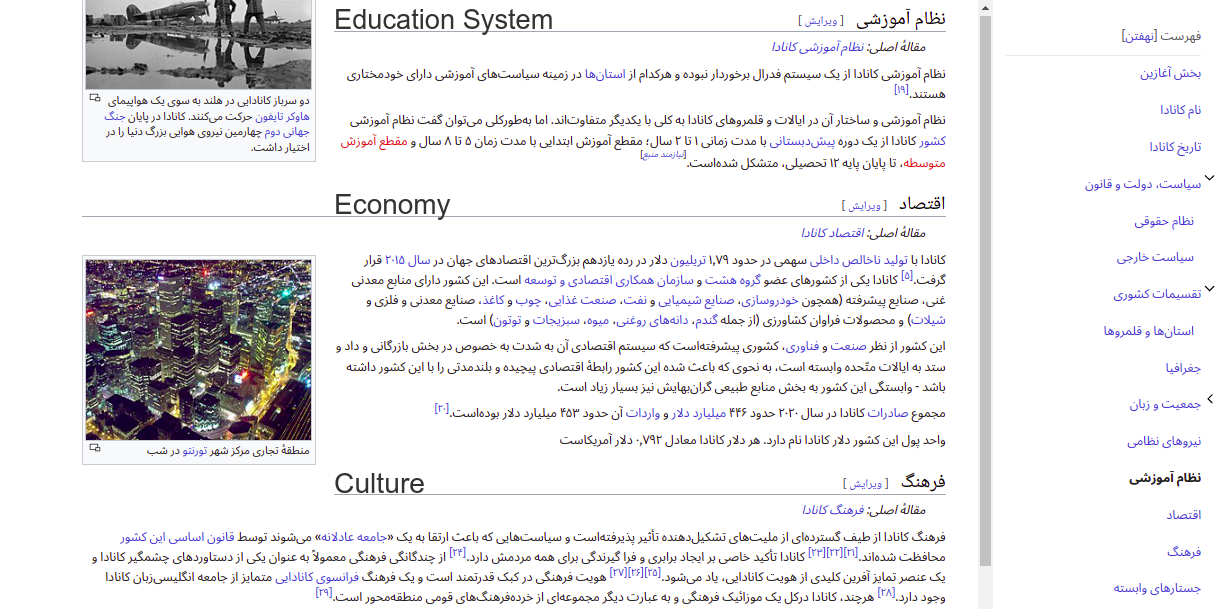}
		\caption
		{
            A segment of the Canada Wikipedia page
		}
		\label{fig:canada}
	\end{center}
\end{figure*}

\end{document}